%% file: main.tex
\documentclass{article}
\PassOptionsToPackage{numbers}{natbib}
 \usepackage[preprint]{neurips_2026}

\usepackage[utf8]{inputenc} 
\usepackage[T1]{fontenc}    
\usepackage{hyperref}       
\usepackage{url}            
\usepackage{booktabs}       
\usepackage{amsfonts}       
\usepackage{nicefrac}       
\usepackage{microtype}      
\usepackage{xcolor}         

\input{preamble}

\title{OpenGaFF: Open-Vocabulary Gaussian Feature Field with Codebook Attention}

\author{\\
    Kunyi Li$^{1,3}$ \quad
    Michael Niemeyer$^{2}$ \quad
    Sen Wang$^{1,3}$\quad
    Stefano Gasperini$^{1,3,4}$ \quad \\
    Nassir Navab$^{1,3}$ \quad 
    Federico Tombari$^{1,2,3}$ \vspace{0.4em} \\
    {\normalsize $^1$Technical University of Munich} \quad 
    {\normalsize $^2$Google} \quad \\
    {\normalsize $^3$Munich Center for Machine Learning}  \quad 
    {\normalsize $^4$Visualais} \\
}

\begin{document}
\maketitle

\vspace{-5mm}
\begin{center}
    \includegraphics[width=0.9\linewidth]{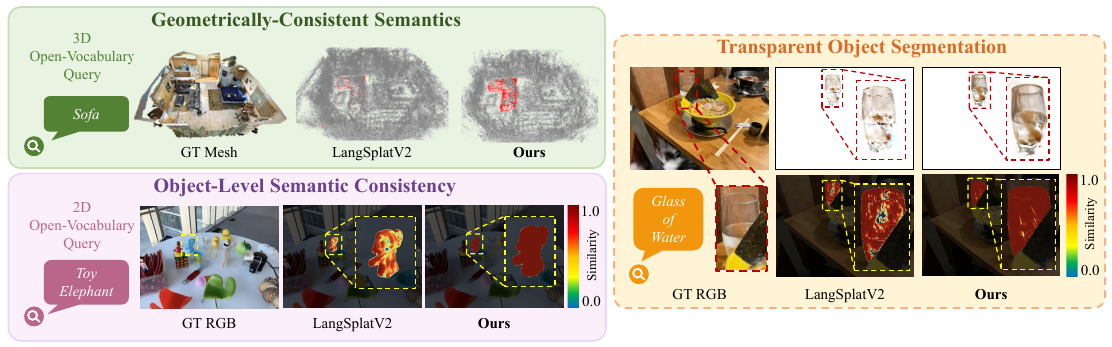}
    \vspace{-2mm}
    \captionof{figure}{\textbf{OpenGaFF} is an open-vocabulary 3D scene understanding method and achieves \textit{precise segmentation} and \textit{consistently high vision-language similarity score} in both 2D and 3D evaluations.}
    \label{fig:teaser}
\end{center}

\input{sec/0_abstract}    
\input{sec/1_intro}
\input{sec/2_related}
\input{sec/3_method}

\input{sec/4_experiment}
\input{sec/5_conclusion}

{
    \small
    \bibliographystyle{unsrtnat}
    \bibliography{main}
    
}


\appendix
\input{sec/X_suppl}



\end{document}

%% file: preamble.tex
\usepackage{tikz}
\usepackage[abs]{overpic}
\usepackage[normalem]{ulem}

\usepackage[symbol]{footmisc}

\usetikzlibrary{fadings,patterns}

\usepackage{blindtext}

\usepackage{colortbl}
\usepackage{graphicx}
\usepackage{subcaption}
\usepackage{arydshln}
\usepackage[acronym]{glossaries}
\usepackage{bm}
\usepackage{multirow} 
\usepackage{booktabs}

\usepackage{wrapfig}

\newcommand{\best}{\cellcolor{tablered}}
\newcommand{\sbest}{\cellcolor{orange}}
\newcommand{\tbest}{\cellcolor{yellow}}

\definecolor{yellow}{rgb}{1, 1, 0.7}
\definecolor{orange}{rgb}{1, 0.85, 0.7}
\definecolor{tablered}{rgb}{1, 0.7, 0.7}
\definecolor{red}{rgb}{1, 0, 0}

\usepackage{xcolor}

\definecolor{blue_text}{RGB}{100, 100, 250}
\definecolor{red_text}{RGB}{180, 0, 0}

\newcommand{\blue}[1]{\textcolor{blue_text}{#1}}
\newcommand{\red}[1]{\textcolor{red_text}{#1}}

%% file: sec/0_abstract.tex
\begin{abstract}
\label{sec:abstract}
Understanding open-vocabulary 3D scenes with Gaussian-based representations remains challenging due to fragmented and spatially inconsistent semantic predictions across multi-view observations. In this paper, we present OpenGaFF, a novel framework for open-vocabulary 3D scene understanding built upon 3D Gaussian Splatting. 
At the core of our method is a Gaussian Feature Field that models semantics as a continuous function of Gaussian geometry and appearance. By explicitly conditioning semantic predictions on geometric structure, this formulation strengthens the coupling between geometry and semantics, leading to improved spatial coherence across similar structures in 3D space. 
To further enforce object-level semantic consistency, we introduce a structured codebook that serves as a set of shared semantic primitives. Furthermore, a codebook-guided attention mechanism is proposed to retrieve language features via similarity matching between query embeddings and learned codebook entries, enabling robust open-vocabulary reasoning while reducing intra-object feature variance.
Extensive experiments on standard 2D and 3D open-vocabulary benchmarks demonstrate that our method consistently outperforms prior approaches, achieving improved segmentation quality, stronger 3D semantic consistency and a semantically interpretable codebook that provides insight into the learned representation. Project page: https://opengaff.github.io/
\end{abstract}

%% file: sec/1_intro.tex
\section{Introduction}
\label{sec:intro}

Understanding 3D scenes and their semantics is a fundamental problem in 3D computer vision, with wide applications in autonomous driving~\cite{wang2024drivedreamer, wang2024driving, xing2025openemma} and robotics~\cite{anderson2018vision, yang2025magma}, among others. A central challenge is to enable machines to perceive complex 3D environments in a manner that is both spatially coherent and semantically meaningful. 
Humans naturally interpret scenes not as collections of independent points, but as structured compositions of objects with consistent semantic identities~\cite{chang2021comprehensive, hughes2024foundations}. Motivated by this observation, an ideal 3D representation should likewise support object-centric and semantically consistent understanding. However, most existing 3D representations~\cite{peng2023openscene, qin2024langsplat, li2025langsplatv2} operate at the level of points or primitives, often resulting in fragmented semantics and limited reasoning capability.

Recent advances in 3D Gaussian Splatting (3DGS)~\cite{kerbl20233d} provide an efficient and explicit 3D representation that enables high-quality rendering, making it a promising foundation for semantic scene understanding. To incorporate semantics, prior works~\cite{li2022language, li2025langsplatv2, wu2024opengaussian, alegret2026gala, liang2026supergseg, wang2026visibility} typically lift 2D semantic features into 3D Gaussians. Despite their effectiveness, these approaches share a fundamental limitation: geometry and semantics are only weakly coupled, leading to semantic representations that lack geometric consistency.

Existing methods fall into three categories.
First, compress-then-decode approaches~\cite{qin2024langsplat, zhou2024feature, peng2026gags} project high-dimensional language features into low-dimensional per-Gaussian embeddings, then rasterize into 2D and decode them at inference time. This irreversible compression discards fine-grained semantic information, limiting open-vocabulary understanding.
Second, feature aggregation methods~\cite{jun2025dr, cheng2024occam, wang2026visibility} directly fuse multi-view features using appearance-based opacity weights. However, such appearance-driven weighting is not always consistent with underlying semantics (e.g., transparent glass), leading to unreliable semantic attribution.
Third, recent codebook-based approaches~\cite{qu2024goi, tian2025ccl, li2025langsplatv2, alegret2026gala} discretize semantics into a set of shared embeddings. While they move towards object-level abstraction, they still assign features in a largely geometry-agnostic manner, leading to spatially inconsistent predictions and fragmented object representations.

In this work, we identify a two-fold design principle: (1) semantics should be closely coupled with geometry, so that spatially similar structures yield consistent predictions; and (2) semantics should be consistent at the object level, so that all regions belonging to the same object share a unified representation.
To this end, we propose OpenGaFF, a novel framework for open-vocabulary 3D scene understanding built upon 3D Gaussian Splatting. At its core, our method introduces a Gaussian Feature Field, which models semantics as a continuous function of Gaussian geometry and appearance. By conditioning semantic predictions on spatial structure, this formulation explicitly couples geometry with semantics.
To further promote object-level consistency, we introduce a structured semantic codebook that serves as a set of shared semantic primitives. Instead of allowing each Gaussian to learn independent language features, we constrain semantics to be composed from this shared codebook. Concretely, regions belonging to the same object are encouraged to map to common codebook entries. This design reduces intra-object feature variation and promotes consistent semantic representation.
Moreover, the codebook is dynamically structured to adapt to scene complexity, enabling a representation that is both compact and expressive.
Overall, our \textbf{contributions} are:
\begin{itemize}
\item We propose a Gaussian Feature Field that conditions on Gaussian geometry and appearance, thereby aligning 3D semantic representation with spatial coherence.
\item We introduce a structured language codebook that enforces object-level consistency by constraining language features to be selected from shared semantic primitives, reducing intra-object variation.
\item We develop a codebook-guided attention mechanism for retrieving high-dimensional language features, enabling consistent and expressive open-vocabulary reasoning in 3D scenes.
\end{itemize}

We evaluate our method on several widely-used 3D open-vocabulary benchmarks. Experimental results demonstrate significant improvements in both 2D and 3D open-vocabulary segmentation as well as localization, with notably enhanced spatial consistency. The code will be publicly released upon acceptance.

%% file: sec/2_related.tex
\section{Related Works}
\label{sec:related}

\subsection{Zero-Shot 2D Scene Understanding}
The remarkable progress of 2D visual foundation models has led to strong performance across a wide range of vision tasks, substantially advancing both perception and reasoning capabilities. These models have also established the foundation for many recent approaches to 3D scene understanding.
DINO~\cite{oquab2024dinov2} learns powerful semantic representations in a self-supervised manner, while Grounding DINO~\cite{liu2024grounding} extends this capability with language guidance for open-vocabulary detection. SAM~\cite{kirillov2023segment} enables prompt-driven segmentation with strong zero-shot generalization. In parallel, CLIP~\cite{radford2021learning} and SigLIP~\cite{tschannen2025siglip} learn aligned image-text embeddings, with SigLIP further improving scalability via a sigmoid-based objective.
Building upon these advances, most existing 3D scene understanding methods~\cite{peng2023openscene, kerr2023lerf, qin2024langsplat} rely on distilling 2D semantic features into 3D representations. While effective, this paradigm inherits the single-view limitation of 2D models, making it difficult to enforce multi-view consistency and achieve holistic 3D understanding.

\subsection{Open-Vocabulary 3D Scene Understanding}
Understanding 3D scenes requires consistent semantic reasoning across multiple views and spatial dimensions. Recent works transfer rich language features from 2D models into 3D representations to enable more human-like perception. OpenScene~\cite{peng2023openscene} distills CLIP features into point clouds for zero-shot segmentation and language querying.
Subsequent approaches~\cite{kerr2023lerf, kim2024garfield, ying2024omniseg3d} embed semantics into neural radiance fields (NeRF) by distilling 2D features, enabling open-vocabulary 3D understanding. However, these methods often suffer from slow training and rendering, suboptimal performance, and high memory consumption due to the cost of volumetric representations.
In contrast, 3DGS provides an explicit and efficient scene representation that is well-suited for real-time 3D understanding. 

\noindent \textbf{Feature Compression Methods.}
LangSplat~\cite{qin2024langsplat} assigns compact semantic features to Gaussians, rasterizes them into 2D feature maps, and uses an MLP-based autoencoder to decode low-dimensional representations into high-dimensional language features. Similarly, Feature3DGS~\cite{zhou2024feature} leverages a CNN to upsample feature dimensions. GAGS~\cite{peng2026gags} further introduces granularity-aware segmentation to enhance supervision signal preprocessing, while also employing MLPs to decode low-dimensional features into language space. However, this compression-and-decode design significantly constrains their ability to achieve comprehensive scene understanding.

\noindent \textbf{Gaussian Clusterization Methods.}
OpenGaussian~\cite{wu2024opengaussian} enables hierarchical 3D clustering and cross-modal 3D–2D feature association, aligning cluster-level scene features with language embeddings. Similarly, SuperGSeg~\cite{liang2026supergseg} aggregates thousands of Gaussians into SuperGaussians with shared language embeddings via an MLP-based cluster updater. However, the clustering in these methods is often complex and lacks semantic consistency, which can lead to the grouping of irrelevant or noisy points, resulting in incorrect clusters.

\noindent \textbf{Feature Aggregation Methods.}
Dr.\ Splat~\cite{jun2025dr}, OccamLGS~\cite{cheng2024occam}, and VALA~\cite{wang2026visibility} adopt a feature aggregation strategy, where multi-view language features are fused in a single forward pass using accumulated opacity weights derived from Gaussian appearance training.
However, these opacity weights are optimized for appearance rendering and may not align with semantic relevance. For example, transparent objects such as glass are often assigned near-zero opacity during appearance learning. As a result, their semantic contributions may be significantly underestimated during feature aggregation.
Furthermore, fusing view-inconsistent language features without filtering or reliability estimation can introduce noise and degrade the representation.

\noindent \textbf{Cdebook-based Methods.}
Instead of compressing language features in dimension, recent methods introduce the codebook to distill multi-view language features into a shared semantic dictionary.
GOI~\cite{qu2024goi} and CCL-LGS~\cite{tian2025ccl} learn a single codebook and use an MLP to predict discrete indices from Gaussian-rendered features. LangSplatV2~\cite{li2025langsplatv2} improves efficiency by applying top-k rasterization on sparse code indices. GALA~\cite{alegret2026gala} further proposes a dual-codebook design with instance–language correspondence, using Gaussian instance features as queries and performing cross-codebook attention to bridge non-semantic and semantic spaces.
Despite its effectiveness, this paradigm suffers from key limitations: a fixed codebook size that cannot adapt to scene complexity, and decoupled geometry and semantics that lead to weak alignment and spatially inconsistent segmentation.

Overall, despite significant progress, existing methods still struggle to produce coherent 3D semantic representations. In contrast, our approach preserves high-dimensional language features, learns geometrically consistent semantic abstractions via structured attention, and introduces an adaptive, semantically grounded codebook that scales with scene complexity.

%% file: sec/3_method.tex
\section{Methods}
\label{sec:method}
\begin{figure*}[t]
\centering
\includegraphics[width=1.0\linewidth]{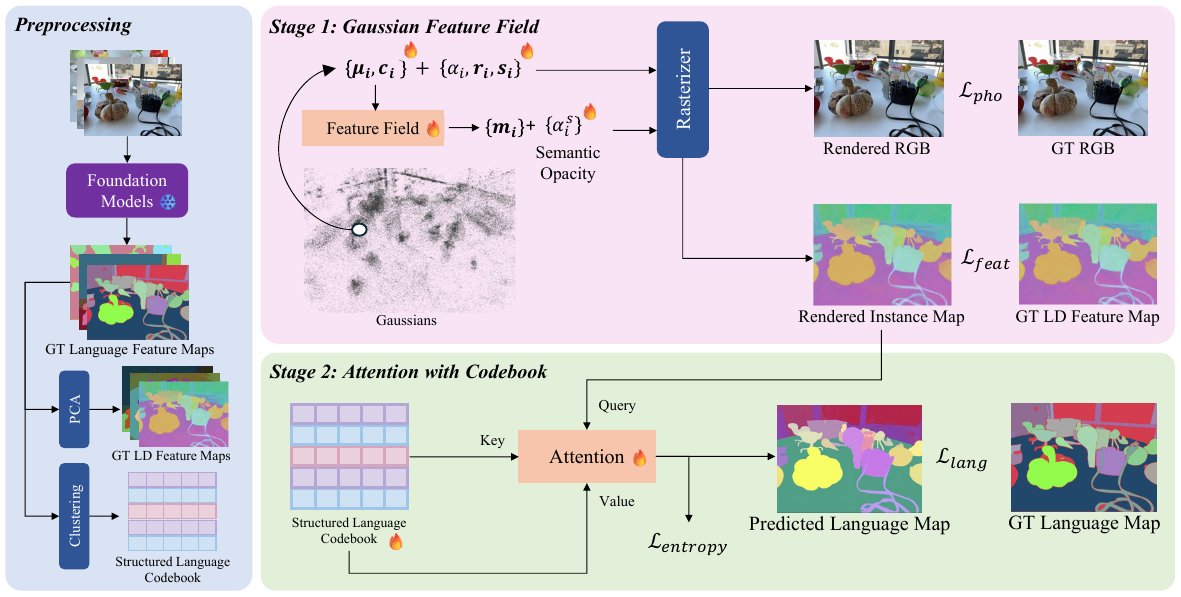}
\caption{\textbf{Overview of OpenGaFF.} We first preprocess RGB images using foundation models to generate ground-truth language features. These features are clustered for structured language codebook initialization, while PCA is applied to per-view feature maps to obtain low-dimensional representations for supervising the Gaussian Feature Field. In Stage 1, the Gaussian Feature Field is trained by 2D feature distillation. In Stage 2, the codebook is trained together with an attention module to capture open vocabulary semantic information.}

\vspace{-5mm}
\label{fig:pipeline} 
\end{figure*}

As illustrated in Figure~\ref{fig:pipeline}, we first preprocess the input to generate the supervision signal, and construct an adaptive structured codebook (Section~\ref{sec:preprocess}). Then, we optimize 3D Gaussian appearance and a geometry-aware feature field jointly (Section~\ref{sec:GFF}). Finally, we train an attention module with the structured codebook to query the language features for visual grounding (Section~\ref{sec:att_codebook}).

\subsection{Preprocessing: LD Feature Map and SL Codebook Initialization}
\label{sec:preprocess}
Given a set of RGB images $\{\mathbf{C}_t\}_{t=1}^{T}$, we preprocess them to construct the supervision signals and initialize the codebook. Following prior works~\cite{qin2024langsplat, li2025langsplatv2, liang2026supergseg, alegret2026gala}, we apply SAM~\cite{kirillov2023segment} to obtain segmentation masks $\mathbb{M} = \{\mathcal{M}_t^{k} \mid k=1,\dots,K_t,\; t=1,\dots,T\}$, and extract vision–language features $\mathbb{F} = \{f_t^{k} \in \mathbb{R}^D \mid k=1,\dots,K_t,\; t=1,\dots,T\}$ using CLIP and reassign them with masks $\mathbb{M}$ as 2D language feature maps $\{\mathbf{L}_t \in \mathbb{R}^{H \times W \times D} \mid t=1,\dots,T\}$.
Unlike existing methods~\cite{li2025langsplatv2, qu2024goi, tian2025ccl, alegret2026gala}, which rely on randomly initialized codebooks with a predefined size, we construct the structured codebook in a data-driven manner. Specifically, we aggregate all CLIP features $\mathbb{F}$ and apply a $k$-means clustering to obtain $N_c$ clusters, where $N_c$ is automatically determined by the structure of the scene. Please refer to the Appendix for how to determine $N_c$. The cluster centroids are then used as the Structured Language (SL) Codebook $S \in \mathbb{R}^{N_c \times D}$, enabling adaptive modeling of the scene complexity.
To provide a more efficient supervision for our Gaussian Feature Field training (Section~\ref{sec:GFF}), we further apply PCA to project all language features into a lower-dimensional (LD) space and generate per-frame LD feature maps $\{\mathbf{M}_t \in \mathbb{R}^{H \times W \times d} \mid t=1,\dots,T\}$. For simplicity, we omit the view index $t$ in the following sections when the context is clear.

\subsection{Gaussian Feature Field}
\label{sec:GFF}
\noindent \textbf{Preliminary.}
3DGS~\cite{kerbl20233d} employs a set of 3D points to render images from given viewpoints. Each 3D Gaussian $\mathcal{G}_{k} = \{{\mu}_{i}, \Sigma_{i}, \alpha_{i}, \mathbf{c}_{i}\}$ consists of a mean position ${\mu}_{i} \in \mathbb{R}^3$, a covariance matrix $\Sigma_{i} = r_{i} s_{i} s_{i}^{\top} r_{i}^{\top}$, $r_{i} \in SO(3), s_{i} \in \mathbb{R}^{3}$, an opacity $\alpha_{i}$ and a color $\mathbf{c}_{i}$.
Given a camera view, through a standard Gaussian Splatting process~\cite{kerbl20233d}, we can render the corresponding color map $\mathbf{\hat{C}}$.

\noindent \textbf{Gaussian Feature Field.}
To incorporate semantic information into 3DGS, prior works~\cite{qin2024langsplat, li2025langsplatv2, alegret2026gala, wu2024opengaussian, qu2024goi, tian2025ccl} assign each Gaussian an additional learnable semantic feature $\mathbf{m}_{i}$, which is then rasterized to produce 2D feature maps from different viewpoints. 
However, this design inherently disentangles geometry and semantics, as each Gaussian learns its feature independently. Although geometrically adjacent Gaussians should correspond to the same object, such spatial coherence is not enforced.
To mitigate this issue, we draw inspiration from NeRF~\cite{mildenhall2021nerf} and introduce a Gaussian Feature Field $\mathcal{F}$. Instead of learning independent per-Gaussian features, we predict the low-dimensional (LD) semantic feature $\mathbf{m}_{i} \in \mathbb{R}^{d}$ conditioned on the Gaussian’s position and color:
\begin{equation}
    \mathbf{m}_{i} = \mathcal{F}(\varphi(\mu_{i}), \mathbf{c}_{i}),
\label{eq:field}
\end{equation}
where $\varphi(\cdot)$ denotes the Fourier positional encoding~\cite{mildenhall2021nerf}, and the Gaussian Feature Field $\mathcal{F}(\cdot)$ is an MLP-based decoder. The feature field implicitly couples geometry and semantics by sharing parameters across Gaussians, encouraging spatially consistent feature representations and improving semantic coherence across the scene. Importantly, the explicit Gaussian representation restricts semantic feature learning to the Gaussian primitives themselves, i.e., learning a mapping from each Gaussian's position to its semantic feature, thereby avoiding unnecessary optimization over free space as in implicit methods~\cite{mildenhall2021nerf, barron2022mip} and improving efficiency.

\noindent \textbf{Separate Semantic Opacity.}
We further introduce a separate semantic opacity $\alpha_i^s$ for each Gaussian to support semantic rendering. This is motivated by the observation that appearance opacity does not necessarily align with semantic importance. We therefore decouple semantic opacity from appearance opacity and rasterize the LD features $\mathbf{m}_{i}$ using $\alpha_i^s$ to obtain the rendered feature map $\mathbf{\hat{M}} \in \mathbb{R}^{H \times W \times d}$. The low-dimensional (LD) semantic supervision is defined as:
\begin{equation}
    \mathcal{L}_{LD}=\big(1-\mathrm{cos}(\mathbf{M}, \mathbf{\hat{M}})\big) + ||\mathbf{M} - \mathbf{\hat{M}}||_1,
\label{eq:ld}
\end{equation}
where $\mathbf{M}$ is the preprocessed ground truth LD feature map.
Therefore, the overall training loss for Stage 1 is: $\mathcal{L}_{stage1}=\mathcal{L}_{pho}+\lambda_{LD}\mathcal{L}_{LD}$, where $\mathcal{L}_{pho}=0.8 \times \| \mathbf{C} - \mathbf{\hat{C}}\|_1 + 0.2 \times SSIM( \mathbf{C}, \mathbf{\hat{C}})$ is the photometric loss~\cite{kerbl20233d}.

\subsection{Attention with Codebook}
\label{sec:att_codebook}
The Gaussian Feature Field distills 2D low-dimensional feature maps $\mathbf{M}$ into a 3D Gaussian representation. However, similar to LangSplat~\cite{qin2024langsplat}, these features are compressed, which inevitably leads to information loss and limits semantic expressiveness. Although LangSplatV2~\cite{li2025langsplatv2} introduces sparse coefficient rendering with a codebook to mitigate this issue, both the codebook size and the number of coefficients are fixed, preventing the representation from adapting to varying scene complexity.  

To address this limitation, we first apply a structured language codebook initialization (Section~\ref{sec:preprocess}), where the codebook is constructed in a data-driven manner. Then we use the codebook as a semantic dictionary and the rendered feature map $\mathbf{\hat{M}}$ as the query. Through an attention mechanism, we retrieve corresponding high-dimensional language features as:
\begin{equation}
A = \mathrm{softmax}\left(\frac{\mathcal{Q}(\mathbf{\hat{M}})\mathcal{K}(S)^\top}{\sqrt{d_{h}}}\right) \in \mathbb{R}^{HW \times N_c}, \ 
\mathbf{\hat{L}} = A\mathcal{V}(S) \in \mathbb{R}^{HW \times D},
\label{eq:attention}
\end{equation}
where $H$ and $W$ are the resolution of the input images, attention weights $A$ denotes the similarities between the queries and the keys, $\mathcal{Q}(\cdot)$ and $\mathcal{K}(\cdot)$ are linear projections, $d_h$ is the hidden dimension, and $\mathcal{V}(\cdot)$ denotes a normalization operator.
We use the preprocessed ground truth language features $\mathbf{L}$ as the language supervision:
\begin{equation}
\mathcal{L}_{lang} = \big(1-\mathrm{cos}(\mathbf{L}, \mathbf{\hat{L}})\big) + \|\mathbf{L} - \mathbf{\hat{L}}\|_1.
\end{equation}
Notably, although training is performed on 2D feature maps, prior work~\cite{alegret2026gala} shows that, due to the linearity of Gaussian rasterization and the approximately linear behavior of attention in the proposed setting, attention and rendering can be treated as commutative: applying attention on rendered features is equivalent to applying it to the underlying 3D Gaussians and then rendering. This property enables learning the codebook using only 2D supervision, while naturally extending to 3D Gaussians at inference for open-vocabulary queries.
To encourage each query to focus on a single codebook entry, we further introduce an entropy regularization term over the attention weights:
\begin{equation}
\mathcal{L}_{entropy} = - \frac{1}{HW} \sum_{i=1}^{HW} \sum_{j=1}^{N_c} A_{ij} \log A_{ij},
\end{equation}
which promotes independence of semantic representations across codebook entries.
Overall, the training loss for Stage 2 is: $\mathcal{L}_{stage2}=\mathcal{L}_{lang}+\lambda_{entropy}\mathcal{L}_{entropy}$.

%% file: sec/4_experiment.tex
\section{Experiments}
\label{sec:experiment}

\subsection{Experimental Setup}
\noindent \textbf{Datasets.}
Following previous methods~\cite{wu2024opengaussian, wang2026visibility, alegret2026gala}, we comprehensively evaluate our method on two real-world datasets: LERF-OVS~\cite{kerr2023lerf} and ScanNet-v2 \cite{dai2017scannet}.

\noindent \textbf{Baselines.} 
We compare our method with a diverse set of 2D and 3D approaches. Specifically, LangSplat~\cite{qin2024langsplat}, GAGS~\cite{peng2026gags}, SuperGSeg~\cite{liang2026supergseg}, and OpenGaussian~\cite{wu2024opengaussian} are feature-based methods; Dr.Splat~\cite{jun2025dr}, OccamLGS~\cite{cheng2024occam}, and VALA~\cite{wang2026visibility} are feature aggregation methods; and GOI~\cite{qu2024goi}, GALA~\cite{alegret2026gala}, and LangSplatV2~\cite{li2025langsplatv2} adopt a codebook-based learning strategy, which we focus on for closer comparison.

\noindent \textbf{Metrics.}
For evaluation, we follow standard practice~\cite{qin2024langsplat, wu2024opengaussian, alegret2026gala, wang2026visibility, jun2025dr}. On LERF-OVS~\cite{kerr2023lerf}, we evaluate on both 2D and 3D. We report open-vocabulary segmentation using mean Intersection-over-Union (mIoU) and object localization accuracy (Acc) on 2D rendered feature map for 2D evaluation. For 3D evaluation, we predict per-Gaussian language features, perform querying in 3D, and then render the selected Gaussians, reporting mIoU and Acc@0.25. On ScanNet-v2~\cite{dai2017scannet}, we evaluate 3D open-vocabulary segmentation with ground truth semantic point cloud, reporting both mIoU and Acc.

\noindent \textbf{Implementation Details.} 
We perform single-GPU training (NVIDIA RTX 4090). For Stage 1, we train 30,000 iterations with $\lambda_{LD}=0.01$ and for Stage 2 we train 10,000 iterations with $\lambda_{entropy}=0.001$. We use $D=512, d=16, d_h=128$. A 3-layers MLP is used as our Feature Field $\mathcal{F}(\cdot)$. For more implementation details, please refer to the Appendix.

\begin{figure*}[t]
\small
\centering
\includegraphics[width=1.\linewidth]{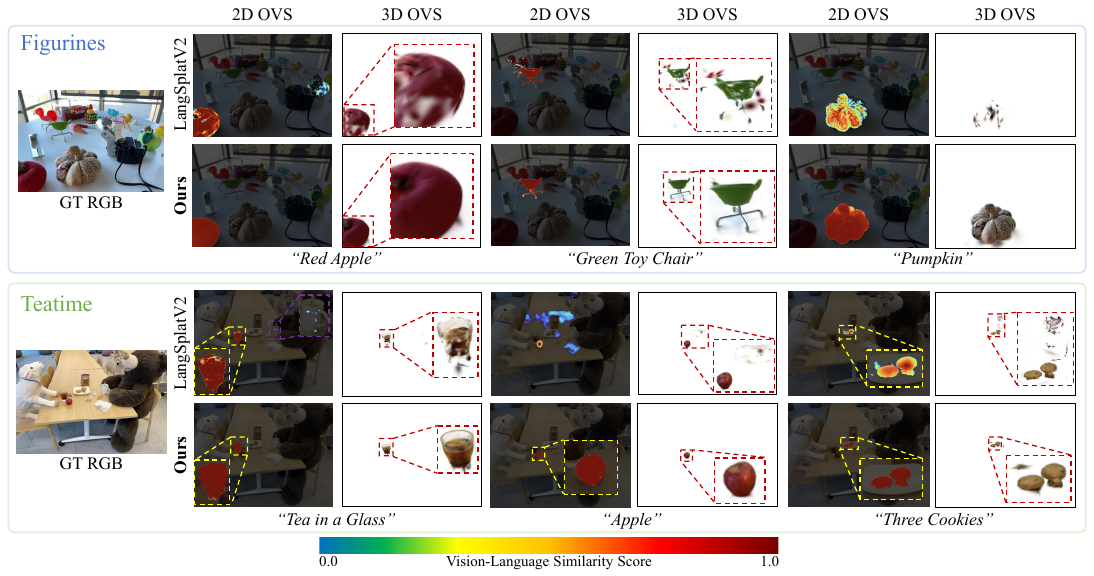}
\caption{\textbf{Qualitative Evaluation of 2D and 3D Open-Vocabulary Query on LERF-OVS~\cite{kerr2023lerf}.} We visualize vision-language similarity as 2D heatmaps. For the shown 3D results, we perform open-vocabulary segmentation directly in 3D and render selected Gaussians. Our method achieve more precise and consistent segmentation in both 2D and 3D.}

\vspace{-5mm}
\label{fig:lerf}

\end{figure*}

\input{tables/lerf-ovs}

\subsection{Evaluation on LERF-OVS}
Table~\ref{table:lerf-ovs} presents open-vocabulary segmentation (OVS) results on the LERF-OVS~\cite{kerr2023lerf} dataset. Since our goal is 3D scene understanding, performing OVS directly in 3D by selecting Gaussians and rendering objects is more faithful than solely 2D OVS on rendered feature maps. For fair comparison, we report both 2D and 3D results. Our method consistently achieves the best performance in both settings.
As shown in Figure~\ref{fig:lerf}, we compare with the state-of-the-art LangSplatV2~\cite{li2025langsplatv2}, which also adopts a codebook-based design. Benefiting from our Gaussian Feature Field (Section~\ref{sec:GFF}), which tightly couples geometry and semantics, our method produces more complete and spatially coherent segmentation results in both 2D and 3D. In 3D OVS, it selects a consistent set of Gaussians for each object, yielding dense and complete renderings. In contrast, LangSplatV2 often produces fragmented or noisy results. Due to weak geometry–semantic coupling, relevant Gaussians may be missed, leading to incomplete shapes (e.g., \textit{"Red Apple"}, \textit{"Green Toy Chair"}, and \textit{"Pumpkin"} in the \textit{Figurines} scene), or it may include semantically unrelated objects (e.g., \textit{"Three Cookies"} in the \textit{Teatime} scene). This arises because their codebook lacks sufficiently disentangled semantic primitives. Further analysis is provided in Section~\ref{sec:discussion}.

\begin{figure*}[t]
\small
\centering
\includegraphics[width=0.9\linewidth]{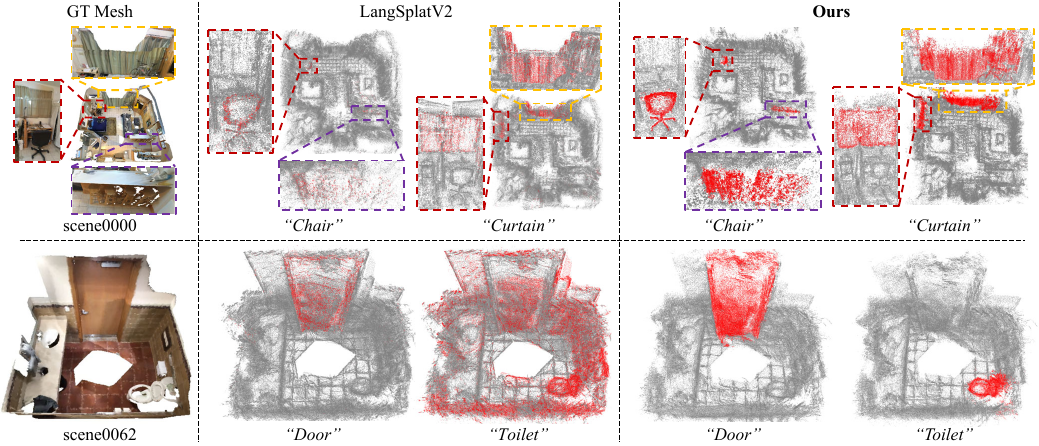}
\caption{\textbf{Qualitative Evaluation of 3D Open-Vocabulary Query on ScanNet-v2~\cite{dai2017scannet}.} 
We highlight the Gaussians corresponding to the query text. Our approach produces more spatially consistent responses while LangSplatV2~\cite{li2025langsplatv2} exhibits significant noise and inconsistent activations.}
\label{fig:scannet}
\vspace{-5mm}
\end{figure*}

\input{tables/scannet}

\subsection{Evaluation on ScanNet-v2}
Table~\ref{tab:scannet} reports 3D open-vocabulary segmentation results on ScanNet-v2~\cite{dai2017scannet}. Our method consistently outperforms existing approaches across all metrics, demonstrating its effectiveness in 3D open-vocabulary understanding.
As shown in Figure~\ref{fig:scannet}, we highlight the point cloud with corresponding queries. LangSplatV2~\cite{li2025langsplatv2} exhibits noticeable noise in predicted semantic point clouds during OVS. While it can partially respond to queries (e.g., \textit{"Chair"} and \textit{"Curtain"} in \textit{scene0000}), many object points are missed, indicating inconsistent per-point semantics. It may also introduce semantic ambiguity, such as jointly activating both the \textit{"Toilet"} and surrounding regions in \textit{scene0062}. In contrast, our method retrieves nearly all object points and produces more complete and coherent segmentation maps.
We attribute these improvements to our Gaussian Feature Field (Section~\ref{sec:GFF}), which tightly couples geometry and semantics in a continuous representation.

\begin{wrapfigure}{r}{0.5\textwidth}
\centering
\vspace{-3mm}
\small
\includegraphics[width=\linewidth]{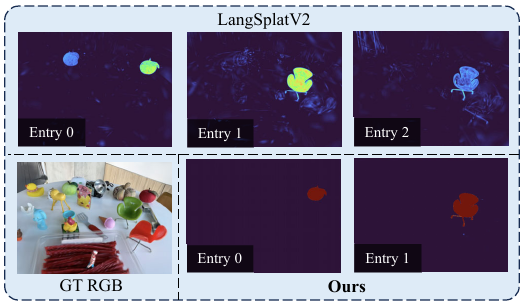}
\caption{\textbf{What Does the Codebook Capture?} LangSplatV2~\cite{li2025langsplatv2} encodes semantics in a distributed manner, where a single entry may represent multiple objects or an object may span multiple entries. In contrast, our method learns disentangled semantic units.}
\vspace{-5mm}
\label{fig:codebook}
\end{wrapfigure}

\subsection{What Does the Codebook Capture?}
\label{sec:discussion}
Recent works~\cite{qu2024goi, alegret2026gala, li2025langsplatv2} have demonstrated the effectiveness of incorporating codebook learning into 3D scene understanding. However, an open question remains: do individual codebook entries correspond to discrete semantic units, or do they merely serve as shared feature bases where semantics emerge only through their combinations?
To investigate this, we visualize per-entry heatmaps on the \textit{Figurines} scene. We set $\lambda_{entropy}=0.01$ in our method for visualization and select several representative entries with the highest responses to the red apple and the green toy chair in Figure~\ref{fig:codebook}. The heatmaps are color-coded from \blue{low response} to \red{high response}.
For LangSplatV2~\cite{li2025langsplatv2}, we observe that multiple entries (Entry 1 and Entry 2 of LangSplatV2) simultaneously respond to the same object (e.g., green toy chair), indicating that semantics are distributed across multiple entries rather than encoded in individual entries. In addition, certain entries (e.g., Entry 0) respond strongly to multiple objects (e.g., red apple and green apple), and even activate on background regions, showing weak semantic specificity and spatial noise. These results suggest that its codebook behaves as an entangled feature base, where semantics emerge only through combinations of entries.
In contrast, our method produces highly localized and semantically consistent responses. Each codebook entry corresponds to a unique and coherent semantic region (e.g., Entry 1 for green toy chair, Entry 0 for red apple) with minimal cross-object activation. This indicates that our codebook learns disentangled semantic units rather than distributed feature bases, improving both interpretability and semantic grounding. We provide further ablation in the Appendix to illustrate how $\lambda_{entropy}$ controls the degree of semantic specificity.

\subsection{Ablation Studies}
\label{sec:ablation}
We conduct comprehensive ablation studies on the \textit{Figurines} and \textit{Ramen} scenes to demonstrate and validate the impact of each contribution.

\begin{figure*}[t]
    \small
    \centering
    \vspace{-5mm}
    \includegraphics[width=1.0\linewidth]{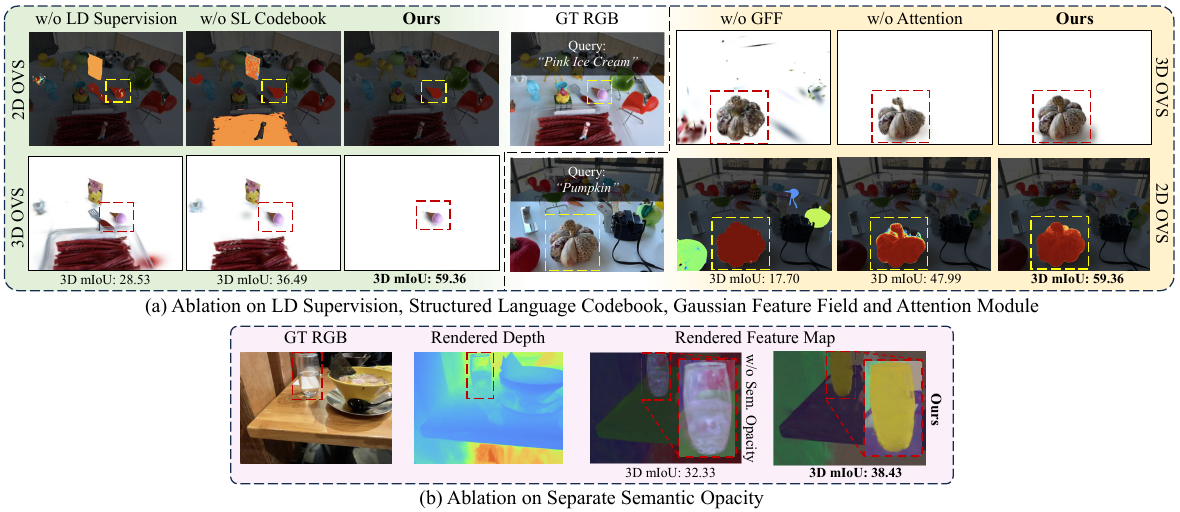}
    \caption{\textbf{Ablation Studies.} We conduct comprehensive ablation studies to demonstrate the effectc of differnet proposed contributions and report the mIoU of 3D OVS on the whole scene.}
    \label{fig:ablation}
    \vspace{-5mm}
\end{figure*}

\noindent \textbf{Ablation on LD feature.} 
As shown in Figure~\ref{fig:ablation}(a), we replace the LD feature supervision (Eq.~\ref{eq:ld}) in Stage 1 with the self-supervised contrastive loss used in prior works~\cite{alegret2026gala, liang2026supergseg}. Without LD semantic supervision, the model can roughly localize the target object \textit{"Pink Ice Cream"}, but also includes irrelevant regions, indicating limited semantic discrimination.
This is because, without LD supervision, the learned low-dimensional features lack explicit semantic guidance and fail to distinguish between objects with similar appearance or geometry.

\noindent \textbf{Ablation on SL codebook.} 
Instead of initializing the structured language codebook via clustering, we adopt random initialization, denoted as "w/o SL Codebook". As shown in Figure~\ref{fig:ablation}(a). Although the predicted segmentation for the query \textit{"Pink Ice Cream"} covers the full object, it also includes substantial irrelevant regions.
This is because random initialization lacks a meaningful semantic prior, preventing codebook entries from efficiently converging to distinct semantic prototypes within the same training budget. Consequently, the learned representations are poorly separated, leading to ambiguous feature matching and causing different objects to be mapped to similar codebook entries.

\noindent \textbf{Ablation on Gaussian Feature Field.}
We disable the Gaussian Feature Field (Eq.~\ref{eq:field}) and instead assign each Gaussian an independent low-dimensional semantic feature following prior work~\cite{alegret2026gala}, denoted as "w/o GFF", as shown in Figure~\ref{fig:ablation}(a). While the model can segment \textit{"Pumpkin"} in both 2D and 3D, they are very noisy, with redundant activations on other regions.
This is because per-Gaussian feature learning ignores geometric consistency on semantics. In contrast, our Gaussian Feature Field explicitly couples semantics with geometry, enabling consistent feature propagation across spatially coherent regions and producing more complete and robust segmentation in both 2D and 3D.

\noindent \textbf{Ablation on Attention Module.}
We replace the codebook-based attention module (Eq.~\ref{eq:attention}) with a 3-layer MLP ( denoted as "w/o Attention" in Figure~\ref{fig:ablation}(a)). This results in incomplete and fragmented segmentation of \textit{"Pumpkin"}, where only partial regions are identified in both 2D and 3D.
This issue stems from multi-view inconsistency, where occlusions and viewpoint changes induce inconsistent supervision signal across an object, leading to intra-object feature variation after 3D distillation. Our codebook-guided attention mitigates this by matching features to shared codebook entries via feature similarity, which is robust to such variations and enforces both semantic and geometric consistency at the object level.
In contrast, the MLP learns independent mappings and tends to overfit view-specific inconsistencies, resulting in non-coherent predictions.

\noindent \textbf{Ablation on Separate Semantic Opacity.}
We conduct this ablation on the \textit{Ramen} scene, which contains a transparent object, i.e., a \textit{"Glass of Water"}. In Figure~\ref{fig:ablation}(b), We visualize the rendered feature maps after PCA. Our method produces a more complete and coherent representation of the glass, whereas the ablated variant ("w/o Sem. Opacity") still exhibits a transparency-induced pattern and fails to form a unified semantic entity.

\input{tables/runtime}

\subsection{Performance Analysis}
We conduct experiments on a single NVIDIA RTX 4090 GPU and report memory, training time, and inference speed on the \textit{Figurines} scene in Table~\ref{tab:runtime}. Both GALA~\cite{alegret2026gala} and LangSplatV2~\cite{li2025langsplatv2} are codebook-based baselines.
Our efficiency gains come from two factors: an implicit Gaussian Feature Field that eliminates per-Gaussian semantic storage, and pixel-sampled training in Stage 2, unlike LangSplatV2~\cite{li2025langsplatv2}, which requires dense rendering.
Dense rendering is unnecessary for semantic supervision, which does not require the fine-grained detail needed for RGB reconstruction.

%% file: tables/lerf-ovs.tex
\begin{table*}[t]
\centering
\small
\caption{\textbf{Quantitative Evaluation of 2D and 3D Open-Vocabulary Query on LERF-OVS~\cite{kerr2023lerf}.} Our method demonstrates strong performance in both 2D and 3D evaluations, especially in 3D tasks, highlighting the advantage of the proposed Gaussian Feature Field.}

\resizebox{0.9\linewidth}{!}{
\begin{tabular}{clcccccccccc}
\toprule
\textbf{Eval.} & \textbf{Method} 
& \multicolumn{2}{c}{\textbf{Mean}} 
& \multicolumn{2}{c}{\textit{Figurines}} 
& \multicolumn{2}{c}{\textit{Teatime}} 
& \multicolumn{2}{c}{\textit{Ramen}} 
& \multicolumn{2}{c}{\textit{Waldo\_kitchen}} \\
\cmidrule(lr){3-4} \cmidrule(lr){5-6} \cmidrule(lr){7-8} \cmidrule(lr){9-10} \cmidrule(lr){11-12}
& 
& mIoU$\uparrow$ & Acc$\uparrow$ 
& mIoU & Acc 
& mIoU & Acc 
& mIoU & Acc 
& mIoU & Acc \\
\midrule

\multirow{7}{*}{2D}
& LangSplat~\cite{qin2024langsplat}    & 51.40 & {84.30\best} & 44.70 & 80.40 & 65.10 & 88.10 & {51.20\tbest} & {73.20\tbest} & 44.50 & {95.50\best}  \\
& GAGS~\cite{peng2026gags}             & 54.12 & 81.66 & 53.59 & 78.57 & 60.29 & 88.14 & 46.81 & 69.01 & 55.80 & {90.91\tbest}  \\
& OccamLGS~\cite{cheng2024occam}       & {61.30\sbest} & 82.50 & {58.60\tbest} & 80.40 & 70.20 & {93.20\sbest} & 51.00 & {74.70\best} & {65.30\sbest} & 81.80   \\
& GOI~\cite{qu2024goi}                 & 42.00 & 59.20 & 23.90 & 44.60 & 55.80 & 67.80 & 33.70 & 56.30 & 54.50 & 68.20  \\
& GALA~\cite{alegret2026gala}          & 55.49 & 73.43 & {59.35\sbest} & {82.14\sbest} & {76.73\best} & 88.14 & 35.13 & 50.70 & 50.75 & 72.73   \\
& LangSplatV2~\cite{li2025langsplatv2} & {59.90\tbest} & {84.10\sbest} & 56.40 & {82.10\tbest} & {72.20\tbest} & {93.20\sbest} & {51.80\sbest} & {74.70\best} & {59.10\tbest} & {95.50\best}   \\
& \textbf{Ours}                        & {64.98\best} & {82.68\tbest} & {64.29\best} & {92.86\best} & {76.09\sbest} & {93.22\best} & {53.78\best} & 62.83 & {65.76\best} & 81.82   \\

\midrule

\multirow{7}{*}{3D}
& OpenGaussian~\cite{wu2024opengaussian}  & 38.36 & 51.43 & 39.29 & 55.36 & {60.44\tbest} & 76.27 & {31.01\tbest} & {42.25\tbest} & 22.70 & 31.82  \\
& SuperGSeg~\cite{liang2026supergseg}     & 35.94 & 52.02 & 43.68 & 60.71 & 55.31 & 77.97 & 18.07 & 23.94 & 26.71 & 45.45  \\
& OccamLGS~\cite{cheng2024occam}          & {47.22\sbest} & {74.84\sbest} & {52.90\tbest} & {78.57\tbest} & {61.02\sbest} & {93.22\best} & {32.01\sbest} & {54.92\sbest} & {42.95\sbest} & {72.72\sbest} \\
& Dr.Splat~\cite{jun2025dr}               & {43.29\tbest} & {64.30\tbest} & {54.42\sbest} & {80.36\sbest} & 57.35 & 77.97 & 24.33 & 35.21 & {37.05\tbest} & {63.64\tbest} \\
& GALA~\cite{alegret2026gala}             & 36.71 & 59.71 & 45.25 & 69.64 & 53.27 & {84.75\tbest} & 17.08 & 25.35 & 31.22 & 59.09 \\
& LangSplatV2~\cite{li2025langsplatv2}   & 35.87 & 55.80 & 45.15 & 67.86 & 49.30 & 79.66 & 19.01 & 21.13 & 30.00 & 54.55   \\
& \textbf{Ours}                           & {54.36\best} & {80.84\best} & {59.36\best} & {92.86\best} & {71.04\best} & {89.83\sbest} & {38.43\best} & {63.38\best} & {48.61\best} & {77.27\best}   \\

\bottomrule
\end{tabular}
}

\label{table:lerf-ovs}
\vspace{-2mm}
\end{table*}

%% file: tables/scannet.tex
\begin{wraptable}{r}{0.6\textwidth}
\centering
\vspace{-5mm}

\caption{\textbf{Quantitative Evaluation of 3D Open-Vocabulary Segmentation on ScanNet-v2~\cite{dai2017scannet}.} Our method achieves the best performance across all evaluation metrics, which we attribute to the proposed Gaussian Feature Field, in contrast to prior methods that rely on per-Gaussian feature learning.}
\resizebox{\linewidth}{!}{
\begin{tabular}{lcccccc}
\toprule
& \multicolumn{2}{c}{19 classes} & \multicolumn{2}{c}{15 classes} & \multicolumn{2}{c}{10 classes} \\
\cmidrule(lr){2-3} \cmidrule(lr){4-5} \cmidrule(lr){6-7}
Method & mIoU $\uparrow$ & mAcc $\uparrow$ & mIoU $\uparrow$ & mAcc $\uparrow$ & mIoU $\uparrow$ & mAcc $\uparrow$ \\
\midrule
LangSplat~\cite{qin2024langsplat} & 2.45 & 8.59 & 3.45 & 13.21 & 6.48 & 21.89 \\
LangSplatV2~\cite{li2025langsplatv2} & 14.75 & 25.47 & 17.09 & 35.68 & 22.83 & 41.52 \\
OpenGaussian~\cite{wu2024opengaussian} & 27.73 & 42.01 & 29.67 & 46.15 & 39.93 & 57.34 \\
Dr.~Splat~\cite{jun2025dr} & 29.31 & 47.68 & 33.25 & {54.33\tbest} & 44.19 & {65.19\tbest} \\
OccamLGS~\cite{cheng2024occam} & {31.93\tbest} & {48.93\tbest} & {34.25\tbest} & 53.71 & {45.16\tbest} & 64.39 \\
VALA~\cite{wang2026visibility} & {32.11\sbest} & {50.05\sbest} & {35.10\sbest} & {54.77\sbest} & {46.21\sbest} & {65.61\sbest} \\
\textbf{Ours} & {36.55\best} & {50.57\best} & {42.78\best} & {72.85\best} & {57.85\best} & {77.93\best} \\
\bottomrule
\end{tabular}
}
\vspace{-3mm}
\label{tab:scannet}
\end{wraptable}

%% file: tables/runtime.tex
\begin{wraptable}{r}{0.48\linewidth}
\centering
\vspace{-5mm}

\caption{\textbf{Performance Analysis.}}
\resizebox{\linewidth}{!}{
\begin{tabular}{lccc}
\hline
Method & Memory & Train & Inference \\
\hline
GALA~\cite{alegret2026gala} & 14 GB & 200 min & 12 fps \\
LangSplatV2~\cite{li2025langsplatv2} & 24 GB & 45 min & \textbf{40 fps} \\
\textbf{Ours} & \textbf{12 GB} & \textbf{15 min} & 35 fps \\
\hline
\end{tabular}
}
\vspace{-3mm}
\label{tab:runtime}
\end{wraptable}

%% file: sec/5_conclusion.tex
\section{Conclusions}
\label{sec:conclusion}

We presented OpenGFF, a framework for open-vocabulary 3D scene understanding based on 3D Gaussian Splatting. Our approach addresses the challenge of learning expressive yet geometrically consistent 3D semantic representations.
To this end, we introduce a Gaussian Feature Field that enables geometry-aware semantic propagation, and a structured codebook with attention-based language retrieval to promote disentangled semantic representations. We further incorporate entropy regularization and semantic opacity to improve spatial consistency and robustness in challenging scenarios such as transparent objects.
Extensive experiments demonstrate consistent improvements in both 2D and 3D open-vocabulary scene understanding, producing more complete and geometry-aligned semantic segmentation results.

%% file: sec/X_suppl.tex
\clearpage
\setcounter{page}{1}

\appendix

\section{Implementation Details}

\noindent \textbf{Preprocessing.}
We follow LangSplat~\cite{qin2024langsplat} to preprocess SAM~\cite{kirillov2023segment} masks and CLIP~\cite{radford2021learning} language features from ground-truth RGB images.

\noindent \textbf{Training.} 
We use GSplat~\cite{ye2025gsplat} as the rasterizer with feature rendering support, following the standard Gaussian Splatting setup. During Stage 1 training, we optimize only the appearance branch for the first 15,000 iterations. After 15,000 iterations, once Gaussian densification is disabled, we jointly optimize both the appearance branch and the proposed Gaussian Feature Field. For Stage 2 training, instead of processing the entire image, we randomly sample 32,776 pixels in each iteration for efficient optimization.

\noindent \textbf{Evaluation.}
We follow the common pratice~\cite{qin2024langsplat, wu2024opengaussian} and further evaluate on MipNeRF360~\cite{barron2022mip} in Section~\ref{sec:mip}.
On LERF-OVS~\cite{kerr2023lerf}, we use reported results from the original papers when available (Table~\ref{table:lerf-ovs}). As some methods are inherently limited to either 2D or 3D and considering computational constraints, we report both 2D and 3D evaluations primarily for codebook-based methods. Since our method is also codebook-based, this ensures a fair and direct comparison.
On ScanNet-v2~\cite{dai2017scannet}, following prior work~\cite{wu2024opengaussian, wang2026visibility}, we evaluate on eight scenes. For each scene, we sample one keyframe every 20 frames for training. The evaluation protocol is adopted from Dr. Splat~\cite{jun2025dr}. We use LangSplatV2's~\cite{li2025langsplatv2} official implementation to perform 3D evaluation on both LERF-OVS and ScanNet-v2.

\section{Structured Language Codebook Initialization}
To initialize the codebook, we perform clustering over all extracted feature vectors from the preprocessed dataset. Specifically, we first collect all feature embeddings $\{\mathbf{L}_i\}_{i=1}^N$, where $\mathbf{L}_i \in \mathbb{R}^{512}$, from all scenes and concatenate them into a single feature matrix.
We then apply PCA to reduce the feature dimensionality from 512 to $d$ (set to 16 in our experiments), obtaining $\{{\mathbf{M}}_i\}_{i=1}^N$, where ${\mathbf{M}}_i \in \mathbb{R}^{H \times W \times d}$. These reduced features are used for Gaussian Feature Field training (Section~\ref{sec:GFF}).

For codebook initialization, we perform $k$-means clustering with cosine similarity. The number of clusters $k$ is selected by searching over a range $[k_{\min}, k_{\max}]$ and choosing the value that maximizes the silhouette score~\cite{shahapure2020cluster}:
\begin{equation}
N_c = \arg\max_k \ \text{Silhouette}(\{\mathbf{L}_i\}, k).
\end{equation}
Given the optimal clustering assignment, we compute the cluster centers as:
\begin{equation}
{S}_j = \frac{1}{|\mathcal{C}_j|} \sum_{\mathbf{L}_i \in \mathcal{C}_j} \mathbf{L}_i,
\end{equation}
where $\mathcal{C}_j$ denotes the set of features assigned to cluster $j$ and $S_j$ is used as the entry of codebook $S \in \mathbb{R}^{N_c \times D}$.

\section{Additional Ablation and Limitation}
\label{sec:supp_discussion}
\subsection{Ablation on Entropy Loss and Limitation}
\begin{figure*}[t]
    \centering
    \includegraphics[width=1.0\linewidth]{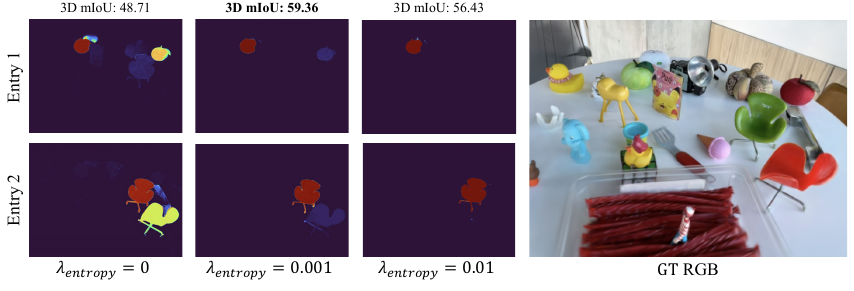}
    \caption{\textbf{Ablation on Entropy Loss.} We report the mIoU of 3D OVS on \textit{Figurines} scene. Larger $\lambda_{entropy}$ values encourage stricter object-level bindings but may over-specialize entries, hurting rare object learning due to limited observations.}
    \label{fig:supp_ablation_entropy}
\end{figure*}

We further present an ablation study on the proposed entropy loss by setting $\lambda_{entropy} = 0, 0.001, 0.01$ and visualizing the corresponding codebook heatmaps in Figure~\ref{fig:supp_ablation_entropy}. When $\lambda_{entropy} = 0$, the codebook captures object-level semantics but each entry often responds to multiple objects with relatively high activation. As $\lambda_{entropy}$ increases to $0.001$, the activations become more concentrated, and each entry tends to focus on a single object, improving semantic exclusivity. When $\lambda_{entropy} = 0.01$, the representation becomes highly specialized, where each entry is almost exclusively associated with one object.
Interestingly, for $\lambda_{entropy}=0$, the multi-object responses are not random but semantically consistent. For example, Entry 1 responds to both red and green apples (shared concept: \textit{apple}) as well as red and green toy chairs (shared concept: \textit{toy chair}), indicating that the codebook can capture compositional semantic factors without entropy regularization. Increasing $\lambda_{entropy}$ gradually suppresses this compositionality and enforces instance-level specialization.
Quantitatively, we find that $\lambda_{entropy} = 0.001$ achieves the best mIoU. We hypothesize that while larger $\lambda_{entropy}$ enforces more exclusive, object-level bindings, it also forces each entry to learn independent semantics, which can be suboptimal for objects that appear infrequently in the training data. Due to limited observations, such objects may not be sufficiently learned, leading to degraded segmentation performance. This highlights a trade-off between semantic compositionality and specialization, which can be effectively controlled by the entropy term.

\subsection{Object-Level Feature Consistency}
\begin{figure*}[t]
    \centering
    \includegraphics[width=1.0\linewidth]{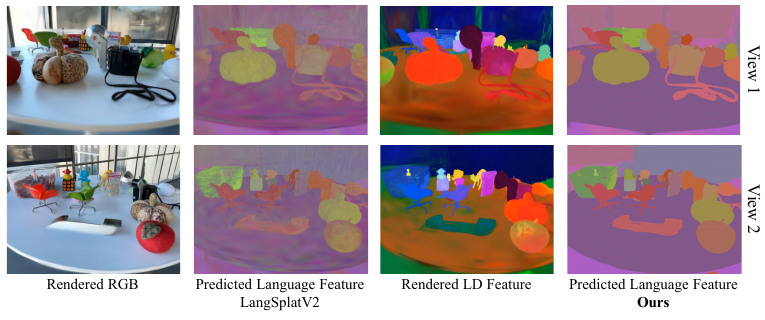}
    \caption{\textbf{Illustration of Object-Level Feature Consistency.}  Compared with the language feature maps predicted by LangSplatV2~\cite{li2025langsplatv2}, ours are more consistent and clearer, demonstrating superior segmentation performance.}
    \label{fig:supp_ablation}
\end{figure*}

In Figure~\ref{fig:supp_ablation}, we further visualize the rendered LD feature map and predicted language feature map.
Our predicted language feature map is highly uniform, where features within each object are nearly identical with negligible variation. In contrast, LangSplatV2's~\cite{li2025langsplatv2} language feature map and our rendered LD feature map exhibit noticeable intra-object variation, particularly for the \textit{"Table"} in the \textit{Figurines} scene.
This improvement stems from two key factors. First, the attention mechanism is computed based on the cosine similarity between query features and codebook entries, which is inherently tolerant to feature variations. Second, the entropy regularization encourages each query to select a single dominant codebook entry as its language feature, further enforcing consistency. Together, these designs lead to more uniform and semantically coherent feature representations. Please refer to supp. mat. for additional video illustration. 

\subsection{Visualization of Codebook Entries}
\begin{figure*}[t]
    \centering
    \includegraphics[width=1.0\linewidth]{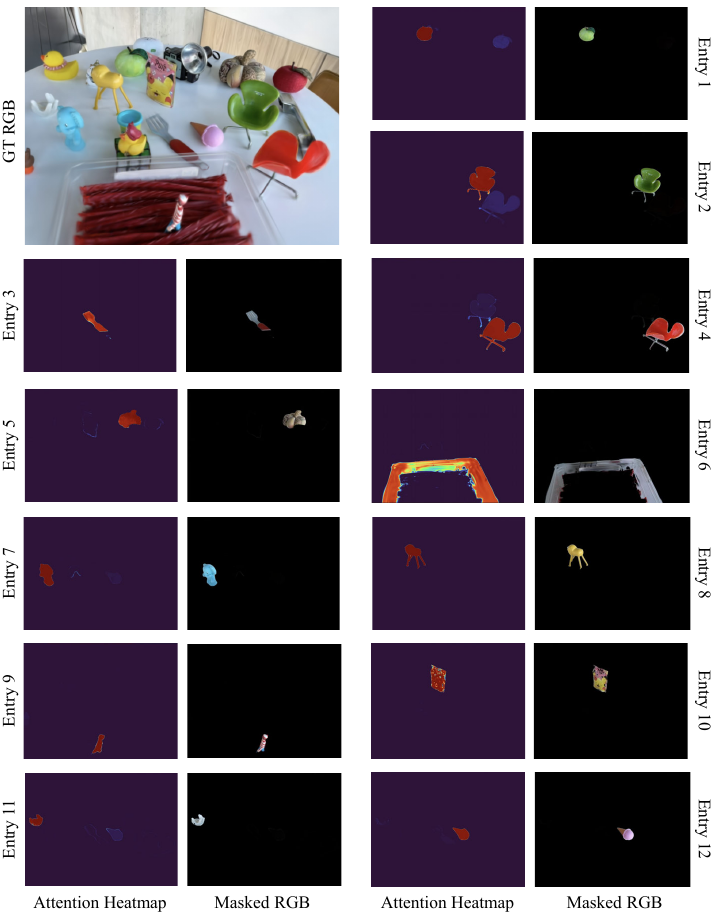}
    \caption{\textbf{Visualization of Codebook Entries.} We present per-entry heatmaps and their corresponding masked RGB images to visualize the regions each codebook entry attends to. These results demonstrate that our codebook effectively captures disentangled and semantically meaningful units.}
    \label{fig:supp_slot}
\end{figure*}

We visualize 12 representative codebook entries in Figure~\ref{fig:supp_slot} to analyze the learned representations. For each entry, we present its attention heatmap and masked RGB image for interpretation.
The results show that each entry consistently attends to a semantically coherent and spatially localized region, typically corresponding to a single object, while different entries exhibit minimal overlap. This suggests that the codebook learns disentangled, object-centric representations rather than performing simple feature aggregation.
Overall, each entry encodes a distinct semantic concept, forming structured and interpretable primitives over the scene.

\section{More Experimental Results}
\subsection{Evaluation on MipNeRF360}
\label{sec:mip}
\begin{figure*}[t]
    \centering
    \includegraphics[width=1.0\linewidth]{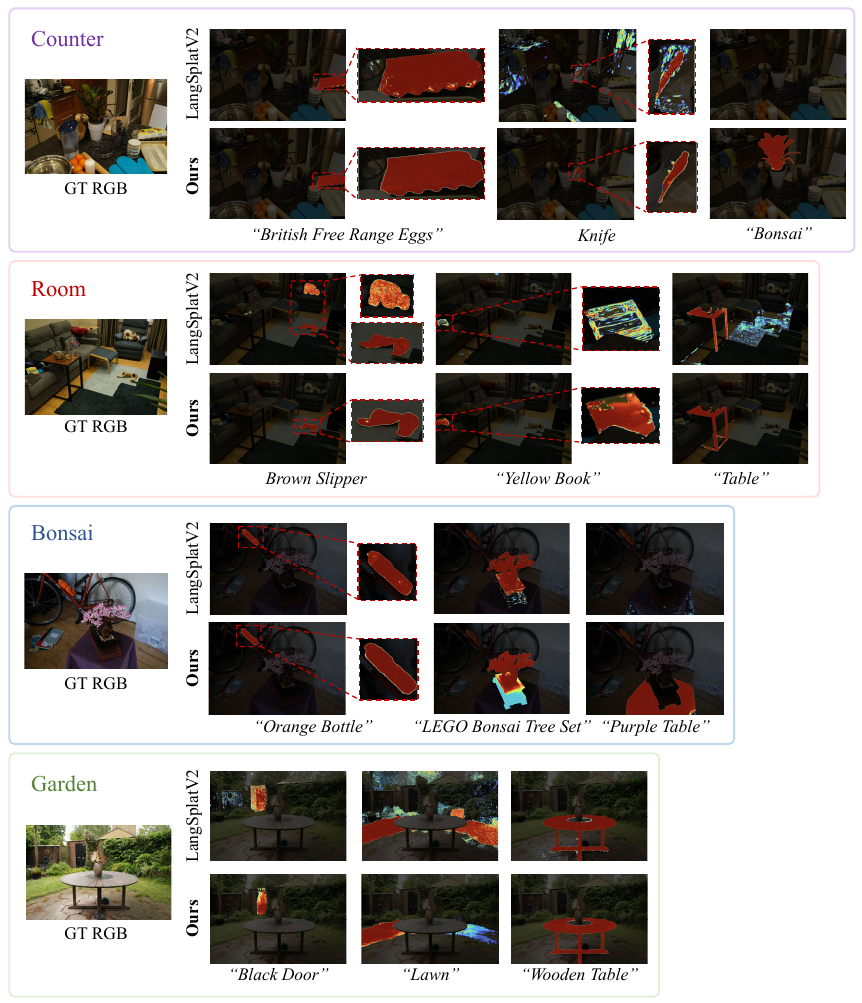}
    \caption{\textbf{Qualitative Evaluation of 2D Open-Vocabulary Segmentation on MipNeRF360~\cite{barron2022mip}.} Our method can predict more precise and consistent segmentation in both 2D and 3D.}
    \label{fig:supp_360}
\end{figure*}

Table~\ref{table:mipnerf360} presents open-vocabulary segmentation (OVS) results on the MipNeRF360~\cite{barron2022mip} dataset. Following prior works~\cite{li2025langsplatv2, peng2026gags}, we report the mIoU of 2D OVS. Our method achieves comparable performance to the state-of-the-art method LangSplatV2~\cite{li2025langsplatv2} in terms of mIoU. We further provide qualitative comparisons with LangSplatV2~\cite{li2025langsplatv2} in Figure~\ref{fig:supp_360}.
Although LangSplatV2~\cite{li2025langsplatv2} achieves slightly higher mIoU scores, we observe that its predictions often exhibit noticeable segmentation noise, where a considerable number of irrelevant regions outside the queried object are also activated. These false responses are typically scattered and can be partially suppressed by the smoothing operations in the evaluation protocol, thus having limited negative impact on the final mIoU.
In contrast, our method produces more spatially precise and semantically consistent segmentation results, with significantly reduced background activation. The predicted regions are better aligned with object boundaries and contain fewer false positives, leading to cleaner and more interpretable outputs in practice. 
We note that the slightly lower mIoU of our method is mainly attributed to challenging cases involving very small objects or objects that appear only sparsely across views. Due to limited observations during training, such instances may not be sufficiently captured by the codebook, leading to occasional misses.
It is also worth noting that prior methods such as LangSplatV2~\cite{li2025langsplatv2} and GAGS~\cite{peng2026gags} are primarily designed and evaluated for 2D open-vocabulary segmentation, with their performance reported mainly on 2D OVS benchmarks. In contrast, our method is designed as a unified framework that supports both 2D and 3D open-vocabulary segmentation. This broader capability introduces additional constraints on the learned representation, which may lead to a slight compromise in 2D metrics. Despite this, our method demonstrates stronger robustness in suppressing spurious activations and yields higher-quality segmentation from both visual and representation perspectives.

\input{tables/mipnerf360}

\subsection{More Evaluation on LERF-OVS}
\begin{figure*}[t]
    \centering
    \includegraphics[width=1.0\linewidth]{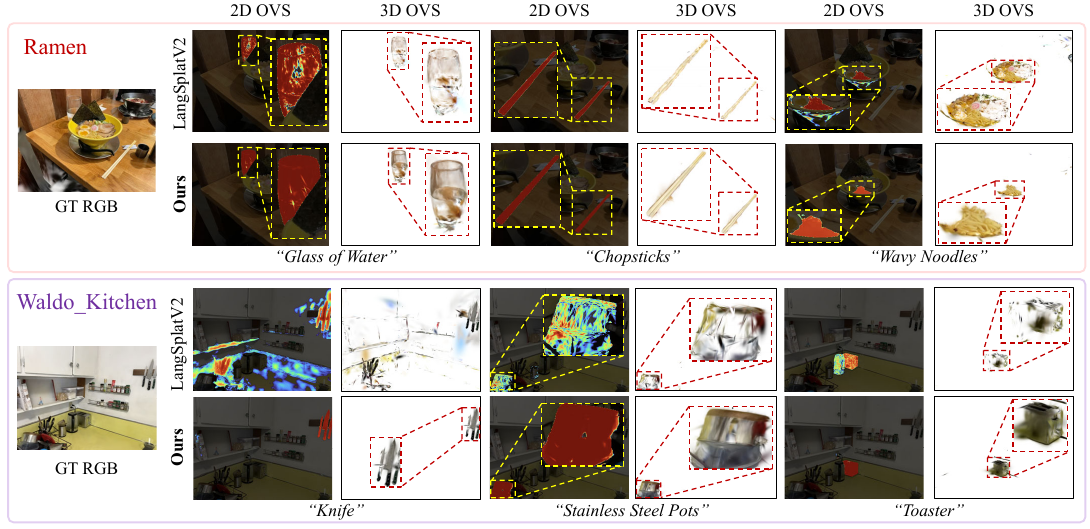}
    \caption{\textbf{Additional Qualitative Evaluation of 2D and 3D Open-Vocabulary Segmentation on LERF-OVS~\cite{kerr2023lerf}.} Our method can predict more precise and consistent segmentation in both 2D and 3D.}
    \label{fig:supp_lerf}
\end{figure*}

We show the 2D and 3D open-vocabulary segmentation results on the \textit{Ramen} and \textit{Waldo\_Kitchen} scenes in Figure~\ref{fig:supp_lerf}, where our method is still able to produce complete and coherent object-level segmentations.

\subsection{More Evaluation on ScanNet-v2}
We visualize the predicted semantic feature point clouds and compare them with the ground-truth semantic point clouds. Compared to LangSplatV2~\cite{li2025langsplatv2}, whose feature point clouds exhibit significant noise with many points misclassified into incorrect categories, leading to poor 3D segmentation quality, our method produces much cleaner and more consistent results. In our case, features within each object are highly uniform, resulting in accurate object-level segmentation. This improvement stems from our Gaussian Feature Field (Section~\ref{sec:GFF}), which effectively couples 3D geometry with semantic representations.



\begin{figure*}[t]
    \centering
    \includegraphics[width=1.0\linewidth]{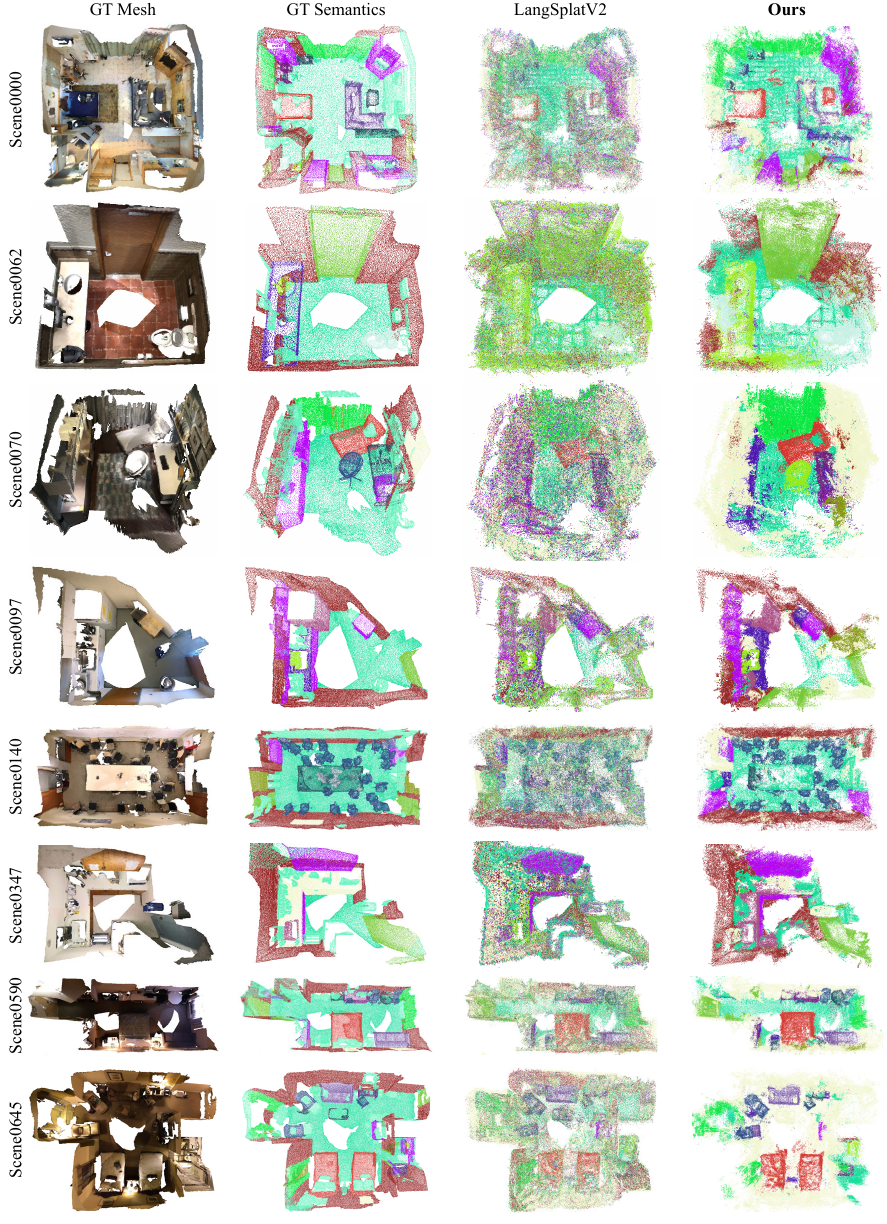}
    \caption{\textbf{Additional Qualitative Evaluation of 3D Open-Vocabulary Segmentation on ScanNet-v2~\cite{dai2017scannet}.} We visualize the language feature point cloud. Ours method can predict clean and more consistent language feature.}
    \label{fig:supp_scannet}
\end{figure*}

%% file: tables/mipnerf360.tex
\begin{table*}[t]
\centering
\small

\caption{\textbf{Quantitative Evaluation of 2D Open-Vocabulary Segmentation on MipNeRF360~\cite{barron2022mip} reported as mean IoU$\uparrow$(\%).} Our method is a unified 2D–3D framework that achieves stronger 3D performance than 2D-focused methods (e.g., LangSplatV2~\cite{li2025langsplatv2}, GAGS~\cite{peng2026gags}), with only a modest trade-off in 2D metrics due to the joint design.}

\resizebox{0.6\linewidth}{!}{
\begin{tabular}{lccccc}
\toprule
\textbf{Method} & \textbf{Mean} & Room & Counter & Garden & Bonsai \\
\midrule
LangSplat~\cite{qin2024langsplat} & 57.3 & 53.2 & \tbest{68.8} & 51.9 & 55.4 \\
LEGaussian~\cite{shi2024language} & 29.1 & 25.5 & 35.3 & 33.2 & 22.3 \\
GOI~\cite{qu2024goi} & 58.5 & 60.3 & 46.6 & \tbest{59.8} & 67.3\\
GAGS~\cite{peng2026gags} & \tbest{64.5} & \best{65.2} & 61.1 & \sbest{61.2} & \sbest{70.5} \\
LangSplatV2 & \best{69.4} & \sbest{64.3} & \sbest{75.1} & \best{65.0} & \best{73.1} \\
Ours & \sbest{65.6} & \tbest{63.0} & \best{75.7} & 55.7 & \tbest{67.8}  \\
\bottomrule
\end{tabular}
}

\label{table:mipnerf360}
\end{table*}

%% file: main.bib
@String(AAAI = {AAAI})

@inproceedings{xing2025openemma,
  title={Openemma: Open-source multimodal model for end-to-end autonomous driving},
  author={Xing, Shuo and Qian, Chengyuan and Wang, Yuping and Hua, Hongyuan and Tian, Kexin and Zhou, Yang and Tu, Zhengzhong},
  booktitle={Proceedings of the Winter Conference on Applications of Computer Vision},
  pages={1001--1009},
  year={2025}
}

@inproceedings{yang2025magma,
  title={Magma: A foundation model for multimodal ai agents},
  author={Yang, Jianwei and Tan, Reuben and Wu, Qianhui and Zheng, Ruijie and Peng, Baolin and Liang, Yongyuan and Gu, Yu and Cai, Mu and Ye, Seonghyeon and Jang, Joel and others},
  booktitle={Proceedings of the computer vision and pattern recognition conference},
  pages={14203--14214},
  year={2025}
}

@inproceedings{anderson2018vision,
  title={Vision-and-language navigation: Interpreting visually-grounded navigation instructions in real environments},
  author={Anderson, Peter and Wu, Qi and Teney, Damien and Bruce, Jake and Johnson, Mark and S{\"u}nderhauf, Niko and Reid, Ian and Gould, Stephen and Van Den Hengel, Anton},
  booktitle={Proceedings of the IEEE conference on computer vision and pattern recognition},
  pages={3674--3683},
  year={2018}
}

@article{kerbl20233d,
  title={3D Gaussian splatting for real-time radiance field rendering.},
  author={Kerbl, Bernhard and Kopanas, Georgios and Leimk{\"u}hler, Thomas and Drettakis, George},
  journal={ACM Trans. Graph.},
  volume={42},
  number={4},
  pages={139--1},
  year={2023}
}

@article{mildenhall2021nerf,
  title={Nerf: Representing scenes as neural radiance fields for view synthesis},
  author={Mildenhall, Ben and Srinivasan, Pratul P and Tancik, Matthew and Barron, Jonathan T and Ramamoorthi, Ravi and Ng, Ren},
  journal={Communications of the ACM},
  volume={65},
  number={1},
  pages={99--106},
  year={2021},
  publisher={ACM New York, NY, USA}
}

@inproceedings{jun2025dr,
  title={Dr. splat: Directly referring 3d gaussian splatting via direct language embedding registration},
  author={Jun-Seong, Kim and Kim, GeonU and Yu-Ji, Kim and Wang, Yu-Chiang Frank and Choe, Jaesung and Oh, Tae-Hyun},
  booktitle={Proceedings of the Computer Vision and Pattern Recognition Conference},
  pages={14137--14146},
  year={2025}
}

@inproceedings{qu2024goi,
  title={Goi: Find 3d gaussians of interest with an optimizable open-vocabulary semantic-space hyperplane},
  author={Qu, Yansong and Dai, Shaohui and Li, Xinyang and Lin, Jianghang and Cao, Liujuan and Zhang, Shengchuan and Ji, Rongrong},
  booktitle={Proceedings of the 32nd ACM international conference on multimedia},
  pages={5328--5337},
  year={2024}
}

@inproceedings{qin2024langsplat,
  title={Langsplat: 3d language gaussian splatting},
  author={Qin, Minghan and Li, Wanhua and Zhou, Jiawei and Wang, Haoqian and Pfister, Hanspeter},
  booktitle={Proceedings of the IEEE/CVF Conference on Computer Vision and Pattern Recognition},
  pages={20051--20060},
  year={2024}
}

@inproceedings{zhou2024feature,
  title={Feature 3dgs: Supercharging 3d gaussian splatting to enable distilled feature fields},
  author={Zhou, Shijie and Chang, Haoran and Jiang, Sicheng and Fan, Zhiwen and Zhu, Zehao and Xu, Dejia and Chari, Pradyumna and You, Suya and Wang, Zhangyang and Kadambi, Achuta},
  booktitle={Proceedings of the IEEE/CVF Conference on Computer Vision and Pattern Recognition},
  pages={21676--21685},
  year={2024}
}

@inproceedings{shi2024language,
  title={Language embedded 3d gaussians for open-vocabulary scene understanding},
  author={Shi, Jin-Chuan and Wang, Miao and Duan, Hao-Bin and Guan, Shao-Hua},
  booktitle={Proceedings of the IEEE/CVF Conference on Computer Vision and Pattern Recognition},
  pages={5333--5343},
  year={2024}
}

@article{wu2024opengaussian,
  title={Opengaussian: Towards point-level 3d gaussian-based open vocabulary understanding},
  author={Wu, Yanmin and Meng, Jiarui and Li, Haijie and Wu, Chenming and Shi, Yahao and Cheng, Xinhua and Zhao, Chen and Feng, Haocheng and Ding, Errui and Wang, Jingdong and others},
  journal={Advances in Neural Information Processing Systems},
  volume={37},
  pages={19114--19138},
  year={2024}
}

@inproceedings{ying2024omniseg3d,
  title={Omniseg3d: Omniversal 3d segmentation via hierarchical contrastive learning},
  author={Ying, Haiyang and Yin, Yixuan and Zhang, Jinzhi and Wang, Fan and Yu, Tao and Huang, Ruqi and Fang, Lu},
  booktitle={Proceedings of the IEEE/CVF Conference on Computer Vision and Pattern Recognition},
  pages={20612--20622},
  year={2024}
}

@inproceedings{barron2022mip,
  title={Mip-nerf 360: Unbounded anti-aliased neural radiance fields},
  author={Barron, Jonathan T and Mildenhall, Ben and Verbin, Dor and Srinivasan, Pratul P and Hedman, Peter},
  booktitle={Proceedings of the IEEE/CVF conference on computer vision and pattern recognition},
  pages={5470--5479},
  year={2022}
}

@inproceedings{kirillov2023segment,
  title={Segment anything},
  author={Kirillov, Alexander and Mintun, Eric and Ravi, Nikhila and Mao, Hanzi and Rolland, Chloe and Gustafson, Laura and Xiao, Tete and Whitehead, Spencer and Berg, Alexander C and Lo, Wan-Yen and others},
  booktitle={Proceedings of the IEEE/CVF international conference on computer vision},
  pages={4015--4026},
  year={2023}
}

@inproceedings{peng2023openscene,
  title={Openscene: 3d scene understanding with open vocabularies},
  author={Peng, Songyou and Genova, Kyle and Jiang, Chiyu and Tagliasacchi, Andrea and Pollefeys, Marc and Funkhouser, Thomas and others},
  booktitle={Proceedings of the IEEE/CVF conference on computer vision and pattern recognition},
  pages={815--824},
  year={2023}
}

@inproceedings{kerr2023lerf,
  title={Lerf: Language embedded radiance fields},
  author={Kerr, Justin and Kim, Chung Min and Goldberg, Ken and Kanazawa, Angjoo and Tancik, Matthew},
  booktitle={Proceedings of the IEEE/CVF international conference on computer vision},
  pages={19729--19739},
  year={2023}
}

@inproceedings{kim2024garfield,
  title={Garfield: Group anything with radiance fields},
  author={Kim, Chung Min and Wu, Mingxuan and Kerr, Justin and Goldberg, Ken and Tancik, Matthew and Kanazawa, Angjoo},
  booktitle={Proceedings of the IEEE/CVF Conference on Computer Vision and Pattern Recognition},
  pages={21530--21539},
  year={2024}
}

@inproceedings{dai2017scannet,
  title={Scannet: Richly-annotated 3d reconstructions of indoor scenes},
  author={Dai, Angela and Chang, Angel X and Savva, Manolis and Halber, Maciej and Funkhouser, Thomas and Nie{\ss}ner, Matthias},
  booktitle={Proceedings of the IEEE conference on computer vision and pattern recognition},
  pages={5828--5839},
  year={2017}
}

@inproceedings{radford2021learning,
  title={Learning transferable visual models from natural language supervision},
  author={Radford, Alec and Kim, Jong Wook and Hallacy, Chris and Ramesh, Aditya and Goh, Gabriel and Agarwal, Sandhini and Sastry, Girish and Askell, Amanda and Mishkin, Pamela and Clark, Jack and others},
  booktitle={International conference on machine learning},
  pages={8748--8763},
  year={2021},
  organization={PmLR}
}

@article{cheng2024occam,
  title={Occam's LGS: An Efficient Approach for Language Gaussian Splatting},
  author={Cheng, Jiahuan and Zaech, Jan-Nico and Van Gool, Luc and Paudel, Danda Pani},
  journal={arXiv preprint arXiv:2412.01807},
  year={2024}
}

@inproceedings{wang2024drivedreamer,
  title={Drivedreamer: Towards real-world-drive world models for autonomous driving},
  author={Wang, Xiaofeng and Zhu, Zheng and Huang, Guan and Chen, Xinze and Zhu, Jiagang and Lu, Jiwen},
  booktitle={European conference on computer vision},
  pages={55--72},
  year={2024},
  organization={Springer}
}

@inproceedings{wang2024driving,
  title={Driving into the future: Multiview visual forecasting and planning with world model for autonomous driving},
  author={Wang, Yuqi and He, Jiawei and Fan, Lue and Li, Hongxin and Chen, Yuntao and Zhang, Zhaoxiang},
  booktitle={Proceedings of the IEEE/CVF Conference on Computer Vision and Pattern Recognition},
  pages={14749--14759},
  year={2024}
}

@article{oquab2024dinov2,
  title={DINOv2: Learning Robust Visual Features without Supervision},
  author={Oquab, Maxime and Darcet, Timoth{\'e}e and Moutakanni, Th{\'e}o and Vo, Huy and Szafraniec, Marc and Khalidov, Vasil and Fernandez, Pierre and Haziza, Daniel and Massa, Francisco and El-Nouby, Alaaeldin and others},
  journal={Transactions on Machine Learning Research Journal},
  year={2024}
}

@inproceedings{liu2024grounding,
  title={Grounding dino: Marrying dino with grounded pre-training for open-set object detection},
  author={Liu, Shilong and Zeng, Zhaoyang and Ren, Tianhe and Li, Feng and Zhang, Hao and Yang, Jie and Jiang, Qing and Li, Chunyuan and Yang, Jianwei and Su, Hang and others},
  booktitle={European conference on computer vision},
  pages={38--55},
  year={2024},
  organization={Springer}
}

@article{tian2025ccl,
  title={CCL-LGS: Contrastive Codebook Learning for 3D Language Gaussian Splatting},
  author={Tian, Lei and Li, Xiaomin and Ma, Liqian and Huang, Hefei and Zheng, Zirui and Yin, Hao and Li, Taiqing and Lu, Huchuan and Jia, Xu},
  journal={arXiv preprint arXiv:2505.20469},
  year={2025}
}

@article{li2022language,
  title={Language-driven semantic segmentation},
  author={Li, Boyi and Weinberger, Kilian Q and Belongie, Serge and Koltun, Vladlen and Ranftl, Ren{\'e}},
  journal={arXiv preprint arXiv:2201.03546},
  year={2022}
}

@article{tschannen2025siglip,
  title={Siglip 2: Multilingual vision-language encoders with improved semantic understanding, localization, and dense features},
  author={Tschannen, Michael and Gritsenko, Alexey and Wang, Xiao and Naeem, Muhammad Ferjad and Alabdulmohsin, Ibrahim and Parthasarathy, Nikhil and Evans, Talfan and Beyer, Lucas and Xia, Ye and Mustafa, Basil and others},
  journal={arXiv preprint arXiv:2502.14786},
  year={2025}
}

@article{liang2026supergseg,
  title={Supergseg: Open-vocabulary 3d segmentation with structured super-gaussians},
  author={Liang, Siyun and Wang, Sen and Li, Kunyi and Niemeyer, Michael and Gasperini, Stefano and Lensch, Hendrik and Navab, Nassir and Tombari, Federico},
  journal={arXiv preprint arXiv:2412.10231},
  year={2024}
}

@article{wang2026visibility,
  title={Visibility-Aware Language Aggregation for Open-Vocabulary Segmentation in 3D Gaussian Splatting},
  author={Wang, Sen and Li, Kunyi and Liang, Siyun and Alegret, Elena and Ma, Jing and Navab, Nassir and Gasperini, Stefano},
  journal={arXiv preprint arXiv:2509.05515},
  year={2025}
}

@article{alegret2026gala,
  title={GALA: Guided Attention with Language Alignment for Open Vocabulary Gaussian Splatting},
  author={Alegret, Elena and Li, Kunyi and Wang, Sen and Liang, Siyun and Niemeyer, Michael and Gasperini, Stefano and Navab, Nassir and Tombari, Federico},
  journal={arXiv preprint arXiv:2508.14278},
  year={2025}
}

@inproceedings{li2025langsplatv2,
  title={LangSplatV2: High-dimensional 3D Language Gaussian Splatting with 450+ FPS},
  author={Li, Wanhua and Zhao, Yujie and Qin, Minghan and Liu, Yang and Cai, Yuanhao and Gan, Chuang},
  booktitle={Annual Conference on Neural Information Processing Systems},
  year={2025}
}

@inproceedings{peng2026gags,
  title={Gags: Granularity-aware feature distillation for language gaussian splatting},
  author={Peng, Yuning and Wang, Haiping and Liu, Yuan and Wen, Chenglu and Dong, Zhen and Yang, Bisheng},
  booktitle={Proceedings of the AAAI Conference on Artificial Intelligence},
  volume={40},
  number={10},
  pages={8376--8384},
  year={2026}
}

@article{ye2025gsplat,
  title={gsplat: An open-source library for Gaussian splatting},
  author={Ye, Vickie and Li, Ruilong and Kerr, Justin and Turkulainen, Matias and Yi, Brent and Pan, Zhuoyang and Seiskari, Otto and Ye, Jianbo and Hu, Jeffrey and Tancik, Matthew and others},
  journal={Journal of Machine Learning Research},
  volume={26},
  number={34},
  pages={1--17},
  year={2025}
}

@inproceedings{shahapure2020cluster,
  title={Cluster quality analysis using silhouette score},
  author={Shahapure, Ketan Rajshekhar and Nicholas, Charles},
  booktitle={2020 IEEE 7th international conference on data science and advanced analytics (DSAA)},
  pages={747--748},
  year={2020},
  organization={IEEE}
}

@article{chang2021comprehensive,
  title={A comprehensive survey of scene graphs: Generation and application},
  author={Chang, Xiaojun and Ren, Pengzhen and Xu, Pengfei and Li, Zhihui and Chen, Xiaojiang and Hauptmann, Alex},
  journal={IEEE Transactions on Pattern Analysis and Machine Intelligence},
  volume={45},
  number={1},
  pages={1--26},
  year={2021},
  publisher={IEEE}
}

@article{hughes2024foundations,
  title={Foundations of spatial perception for robotics: Hierarchical representations and real-time systems},
  author={Hughes, Nathan and Chang, Yun and Hu, Siyi and Talak, Rajat and Abdulhai, Rumaia and Strader, Jared and Carlone, Luca},
  journal={The International Journal of Robotics Research},
  volume={43},
  number={10},
  pages={1457--1505},
  year={2024},
  publisher={SAGE Publications Sage UK: London, England}
}
